\documentclass{article}

\usepackage{microtype}
\usepackage{graphicx}
\usepackage{subfigure}
\usepackage{booktabs} % for professional tables

\usepackage{hyperref}

\usepackage[accepted]{icml2024}

\usepackage{amsmath}
\usepackage{amssymb}
\usepackage{mathtools}
\usepackage{amsthm}

\usepackage{multirow}
\usepackage{array}

\usepackage[capitalize,noabbrev]{cleveref}

\theoremstyle{plain}

\theoremstyle{definition}

\theoremstyle{remark}

\usepackage[textsize=tiny]{todonotes}

\icmltitlerunning{ATraDiff: Accelerating Online Reinforcement Learning with Imaginary Trajectories}

\begin{document}

\twocolumn[
\icmltitle{ATraDiff: Accelerating Online Reinforcement Learning with Imaginary Trajectories}

% It is OKAY to include author information, even for blind
% submissions: the style file will automatically remove it for you
% unless you've provided the [accepted] option to the icml2024
% package.

% List of affiliations: The first argument should be a (short)
% identifier you will use later to specify author affiliations
% Academic affiliations should list Department, University, City, Region, Country
% Industry affiliations should list Company, City, Region, Country

% You can specify symbols, otherwise they are numbered in order.
% Ideally, you should not use this facility. Affiliations will be numbered
% in order of appearance and this is the preferred way.
% \icmlsetsymbol{equal}{*}

\begin{icmlauthorlist}
\icmlauthor{Qianlan Yang}{UIUC}
\icmlauthor{Yu-Xiong Wang}{UIUC}
%\icmlauthor{}{sch}
%\icmlauthor{}{sch}
\end{icmlauthorlist}

\icmlaffiliation{UIUC}{Department of Computer Science, University of Illinois Urbana-Champaign, Urbana, Illinois, USA}
% \icmlaffiliation{comp}{Company Name, Location, Country}
% \icmlaffiliation{sch}{School of ZZZ, Institute of WWW, Location, Country}

\icmlcorrespondingauthor{Qianlan Yang}{qianlan2@illinois.edu}
% \icmlcorrespondingauthor{Firstname2 Lastname2}{first2.last2@www.uk}

% You may provide any keywords that you
% find helpful for describing your paper; these are used to populate
% the "keywords" metadata in the PDF but will not be shown in the document
\icmlkeywords{Machine Learning, ICML, Reinforcement Learning, Generative Model}

\vskip 0.3in
]

% this must go after the closing bracket ] following \twocolumn[ ...

% This command actually creates the footnote in the first column
% listing the affiliations and the copyright notice.
% The command takes one argument, which is text to display at the start of the footnote.
% The \icmlEqualContribution command is standard text for equal contribution.
% Remove it (just {}) if you do not need this facility.

\printAffiliationsAndNotice{}  % leave blank if no need to mention equal contribution
% \printAffiliationsAndNotice{\icmlEqualContribution} % otherwise use the standard text.

\begin{abstract}
Training autonomous agents with sparse rewards is a long-standing problem in online reinforcement learning (RL), due to low data efficiency. Prior work overcomes this challenge by extracting useful knowledge from offline data, often accomplished through the learning of action distribution from offline data and utilizing the learned distribution to facilitate online RL. However, since the offline data are given and fixed, the extracted knowledge is inherently limited, making it difficult to generalize to new tasks. We propose a novel approach that leverages offline data to learn a generative diffusion model, coined as \emph{Adaptive Trajectory Diffuser (ATraDiff)}. This model generates synthetic trajectories, serving as a form of data augmentation and consequently enhancing the performance of online RL methods. The key strength of our diffuser lies in its adaptability, allowing it to effectively handle varying trajectory lengths and mitigate distribution shifts between online and offline data. Because of its simplicity, ATraDiff \emph{seamlessly integrates with a wide spectrum of RL methods}. Empirical evaluation shows that ATraDiff consistently achieves state-of-the-art performance across a variety of environments, with particularly pronounced improvements in complicated settings. Our code and demo video are available at \url{https://atradiff.github.io}.
\end{abstract}

\section{Introduction}

Deep reinforcement learning (RL) has shown great promise in various applications, such as autonomous driving~\citep{wang2019deep}, chip design~\citep{Mirhoseini2021AGP}, and energy optimization~\citep{SPECHT2023100215}. Despite its impressive performance, RL often requires extensive online interactions with the environment, which can be prohibitively costly in practice. Such downside of RL is aggravated by the fact that in real-world scenarios, environments are often characterized by sparse rewards, which further necessitates an exceptionally large number of samples for effective exploration. For example, when manipulating a robotic arm to move an item, oftentimes the only reward feedback given is at the success moment of the task, which may take hundreds of steps to obtain. Consequently, a persistent challenge in RL is addressing the high sample costs, particularly in contexts with sparse rewards. 
\begin{figure}[t]

    \begin{minipage}{0.49\textwidth}
        \centering
        \includegraphics[height=3.0cm]{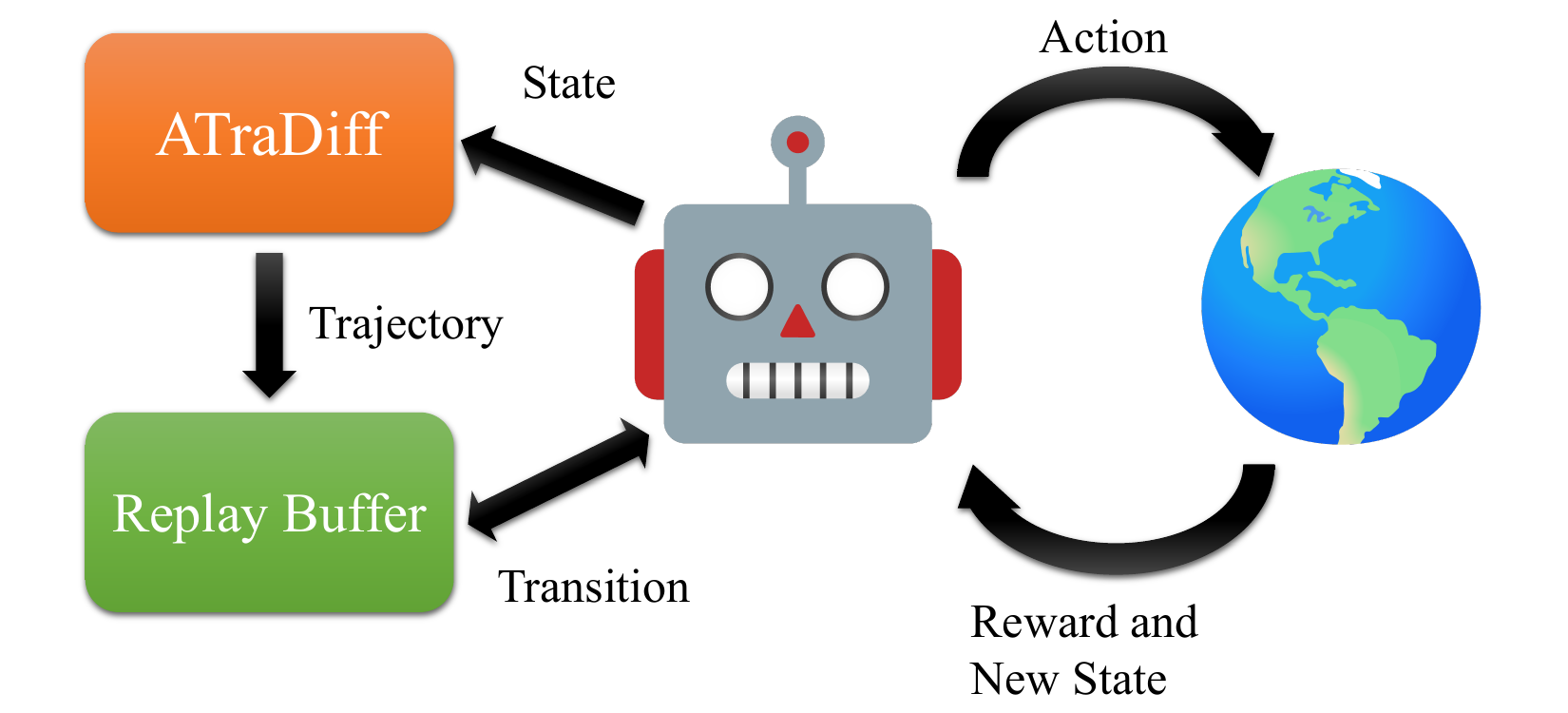}
    \end{minipage}
    \begin{minipage}{0.47\textwidth}
        \centering
        \includegraphics[height=4.2cm]{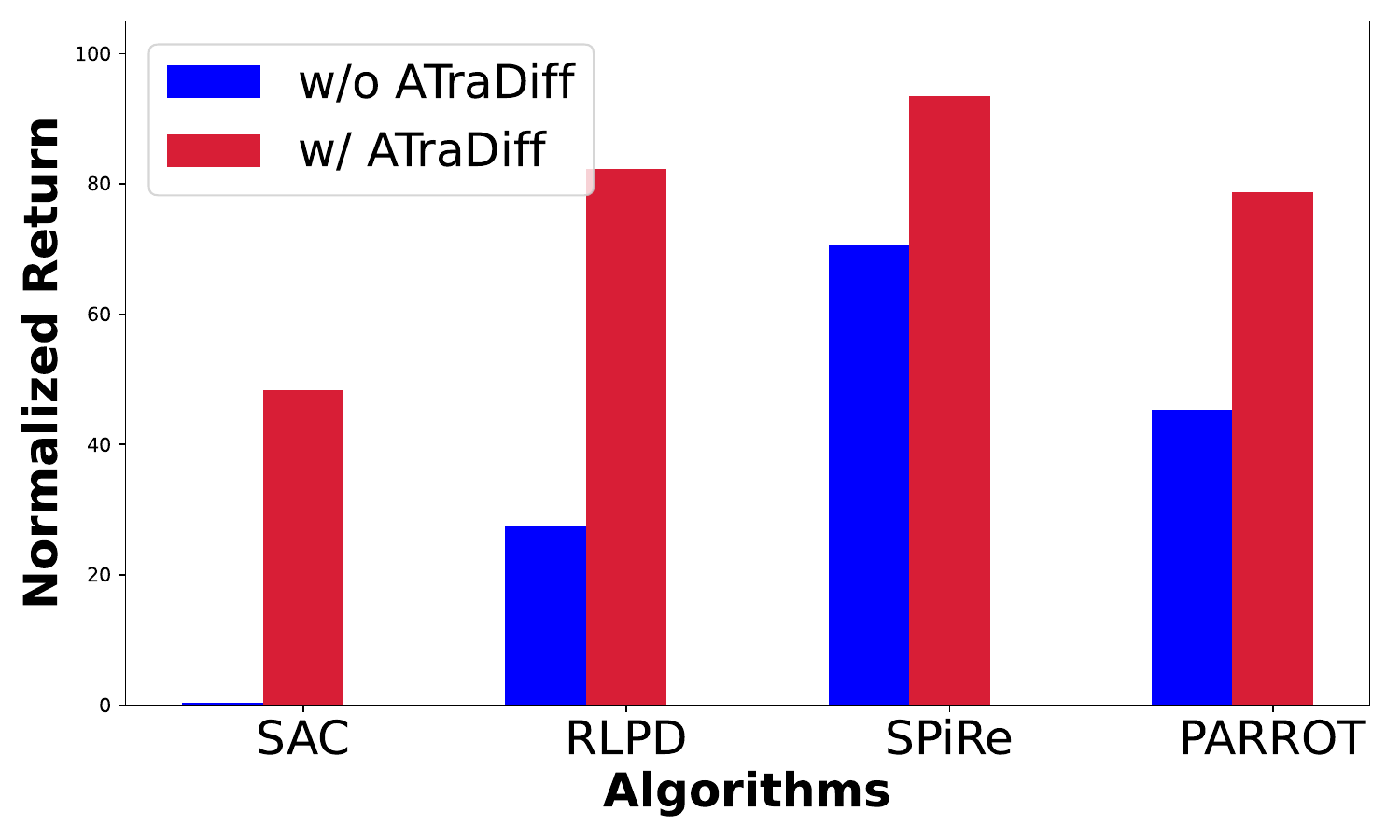}
    \end{minipage}
    \label{figure:illus}
    \caption{Illustration and performance showcase of our ATraDiff. ATraDiff can seamlessly integrate with a wide range of RL methods and consistently improve their performance, by augmenting the replay buffer with synthesized trajectories. \textbf{Top:} Overview of online RL with ATraDiff. \textbf{Bottom:} Performance comparison of RL methods with and without ATraDiff in D4RL Kitchen.}
\end{figure}

One prevalent solution to this challenge is leveraging offline data, by directly learning policies from offline data~\citep{iql, ball2023efficient} or extracting experiences that can enhance online training~\citep{pertsch2020accelerating}, especially in the context of exploration. However, such solutions can only extract limited knowledge as the offline data are given and fixed, and thus are difficult to generalize to new tasks. In contrast to prior work, this paper takes a different perspective inspired by the recent advances in generative modeling -- \emph{can we harness modern generative models, such as diffusion models, trained on offline data and synthesize useful data that facilitate online RL?}

Indeed, diffusion models have emerged as powerful deep generative models, demonstrating impressive capabilities in data synthesis across vision and language applications~\citep{ho2020denoising, gong2023diffuseq, li2022diffusionlm}. Nevertheless, their investigation in RL has been relatively limited. Studies in offline RL mainly tap into the generative capabilities of diffusion models to enhance long-term planning~\citep{janner2022planning, ajay2023conditional} and amplify policy expressiveness~\citep{DBLP:conf/iclr/WangHZ23, DBLP:conf/iclr/Chen0Y0023}. Recently, SynthER~\citep{lu2023synthetic} is introduced to utilize diffusion models for RL data augmentation, by upsampling the data with a diffusion model trained on an offline dataset. However, such an approach primarily focuses on generating \emph{transitions} rather than \emph{complete trajectories}. This under-utilizes the generative potential of diffusion models and limits the benefits of augmented data for RL. In contrast, full trajectories offer a more comprehensive source of information, enabling RL agents to better learn from past experiences.

To overcome these issues, we propose \emph{Adaptive Trajectory Diffuser (ATraDiff)}, a novel method designed to synthesize full trajectories for online RL. As depicted in Figure~\ref{figure:illus}, our approach trains a diffusion model using offline data, which then synthesizes complete trajectories conditioned on the current state. By employing this diffusion model to produce additional trajectories, we aim to significantly accelerate the online RL process. Notably, because of its simplicity in augmenting the replay buffer by adding useful data, ATraDiff \emph{seamlessly integrates with a wide range of RL methods} and \emph{consistently} elevates their performance.

The key property of our diffuser lies in its \emph{adaptability} to effectively handle varying trajectory lengths and address the distribution shifts between online and offline data. Unlike generating transitions, managing the uncertainty in task lengths presents a significant new challenge in trajectory generation. While longer trajectories can potentially lead to improved performance, excessive or redundant segments may be detrimental. Ideally, we aim for a generation with a precise trajectory length. To this end, we introduce a simple yet effective \emph{coarse-to-precise} strategy: initially, we train multiple diffusion models with varying generation lengths. Prior to actual generation, we assess the required length and subsequently prune any redundant segments. In dealing with the distribution shift between offline data and online evaluation tasks, we design our diffuser to be adaptable throughout the RL process. This adaptability includes the capability to select more informative samples through the use of an \emph{importance indicator}, while also mitigating catastrophic forgetting during adaptation.

\textbf{Our contributions} are three-fold.
\textbf{(i)} We propose ATraDiff, a novel diffusion-based approach that leverages offline data to generate full synthetic trajectories and enhance the performance of online RL methods. ATraDiff is general and can be seamlessly applied to accelerate \emph{any} online RL algorithm with a replay buffer, as well as being effective in offline environments.
\textbf{(ii)} We introduce a simple yet effective coarse-to-precise strategy that ensures generated trajectories precisely align with the length required for evaluation tasks.
\textbf{(iii)} We devise an online adaptation mechanism that successfully addresses challenges stemming from data distribution shifts.
Empirical evaluation shows that ATraDiff consistently achieves state-of-the-art performance across a variety of \emph{online, offline-to-online, and offline} environments, with particularly large improvements in complicated settings.

\section{Related work}

\textbf{Offline pretraining for online RL.} Leveraging prior experiences to expedite online learning for subsequent tasks has been a persistent challenge~\citep{nair2021awac}. Past research has proposed numerous solutions to tackle this issue. Some studies suggest treating offline data similarly to data collected online. Representative approaches employ offline data to initialize a replay buffer~\citep{ddpgfd,dqfd}. Meanwhile, others advocate for a balanced sampling strategy, drawing from both offline and online sources~\citep{nair2018overcoming,qt-opt,modem,zhang2023policy, ball2023efficient}.

One prevalent strategy is to establish a behavior prior, which captures the action distribution in prior experiences to mitigate overestimation for actions outside the training data distribution~\citep{singh2020parrot, siegel2020doing}. An alternative strategy involves extracting skills from offline data and adapting them to new tasks~\citep{gupta2019relay, merel2019neural, kipf2019compile, whitney2020dynamicsaware, pertsch2020accelerating}. These studies typically represent the acquired skills within an embedding space and then train a policy to select the most appropriate skills based on the current state. In contrast, ATraDiff synthesizes the complete trajectory based on the current state and augments the replay buffer with these additional data. Our approach offers broader applicability across a diverse set of RL methodologies.

\textbf{Diffusion models in RL.} The focus of employing diffusion models in RL has primarily centered on enhancing long-term planning and amplifying policy expressiveness. For instance, Diffuser~\citep{janner2022planning} constructs a full trajectory of transitions, through conditioned sampling guided by higher rewards and goal-oriented navigation. This leverages the diffusion model's capability in generating extensive trajectories, addressing challenges such as long horizons and sparse rewards in RL planning. Other studies~\citep{ajay2023conditional, du2023learning, DBLP:journals/corr/abs-2305-18459} have adopted this paradigm, particularly in the context of visual data. Another notable work~\citep{DBLP:conf/iclr/PearceRKBSGMTMH23} suggests a diffusion-based method for generating full trajectories by imitating human behavior. Different from prior work, our approach is oriented towards synthesizing trajectories in evaluation scenarios that may differ from the training scenarios.

\textbf{Data augmentation in RL.} Data augmentation is a common technique that has demonstrated effectiveness in RL. Previous methods~\citep{DBLP:conf/iclr/YaratsKF21, DBLP:conf/nips/LaskinLSPAS20, s4rl} typically focus on perturbing original observation data in visual-based RL, such as adding noise and applying random translation. This enables agents to learn from multiple views of the same observation and increase their robustness. Recent efforts have focused on upsampling the replay buffer with the diffusion model. A closely related study, SynthER~\citep{lu2023synthetic}, generates transitions to augment the replay buffer via a diffusion model. However, our ATraDiff \emph{operates at the trajectory level, can employ a visual-based diffusion model, and has the capability to synthesize training data through state- and task-conditioned generation}. Furthermore, our work primarily focuses on leveraging offline data to enhance online RL performance. In contrast, SynthER aims to upsample the offline dataset using its data synthesizer. Our setting thus presents more complex challenges, involving dynamic generation and managing the distribution shift between offline and online data. Another work~\citep{DBLP:journals/corr/abs-2305-18459} investigates training a diffusion model to enhance multi-task learning, whereas we develop a general method to improve any RL method.

\vspace{0.15cm}
\section{Background}

\textbf{MDP.} In this paper, we consider sequential decision-making tasks that can be modeled as a Markov Decision Process (MDP) defined as $\mathcal{M} = \langle S, A, T, R, \gamma \rangle$, where $S$ is the set of states, $A$ is the set of actions, and $\gamma\in[0,1)$ is the discount factor. $T(s'|s, a)$ and $R(s, a)$ represent the dynamics and reward functions, respectively. At each stage $t$, the agent takes an action $a \in A$, which leads to a next state $s'$ according to the transition function $T(s'|s, a)$ and an immediate reward $R(s,a)$. A trajectory of such a task is defined as a sequence composed of states and actions given by $(s_1, a_1, s_2, a_2, \dots, s_t, a_t)$, where $s_t$ and $a_t$ denote the state and action at time-step $t$, respectively.

\textbf{Diffusion models.} Diffusion probabilistic models pose the data-generating process as an
iterative denoising procedure $p_{\theta}(\tau^{i-1}|\tau^i)$. This denoising is the reverse of a forward diffusion process $q(\tau^i|\tau^{i-1})$ that slowly corrupts the structure in data by adding noise in $N$ steps, where $\tau^i$ is the $i$-step of the denoising process. The data distribution induced by the model is given by:
$$
p_{\theta}(\tau^0) = \int p(\tau^N) \prod_{i=1}^N p_{\theta}(\tau^{i-1}|\tau^i) d\tau^{1:N},
$$
where $p(\tau^N)$ is a standard Gaussian prior and $p(\tau^0)$ denotes
noiseless data. Parameters $\theta$ are optimized by minimizing a variational bound on the negative log-likelihood of the reverse process: $ \theta^* = \arg \min_\theta -\mathbb{E}_{\tau^0} [\log p_{\theta}(\tau^0) ]$. The reverse process is often parameterized as Gaussian with fixed timestep-dependent covariances:
$$
p_{\theta}(\tau^{i-1}|\tau^i) = \mathcal{N}(\tau^{i-1}|\mu_{\theta}(\tau^i,  i), \Sigma^i).
$$
\textbf{Replay buffers.} Many algorithms in RL employ replay buffers, denoted as $D$, to retain trajectories derived from executing a sample policy within an environment parameterized by MDP. Throughout the training process, these replay buffers are accessed to extract samples (transitions or trajectories) for updating the learned execution policy.

\section{Method}

\begin{figure*}[t]
    \centering
    \includegraphics[width=.9\linewidth]{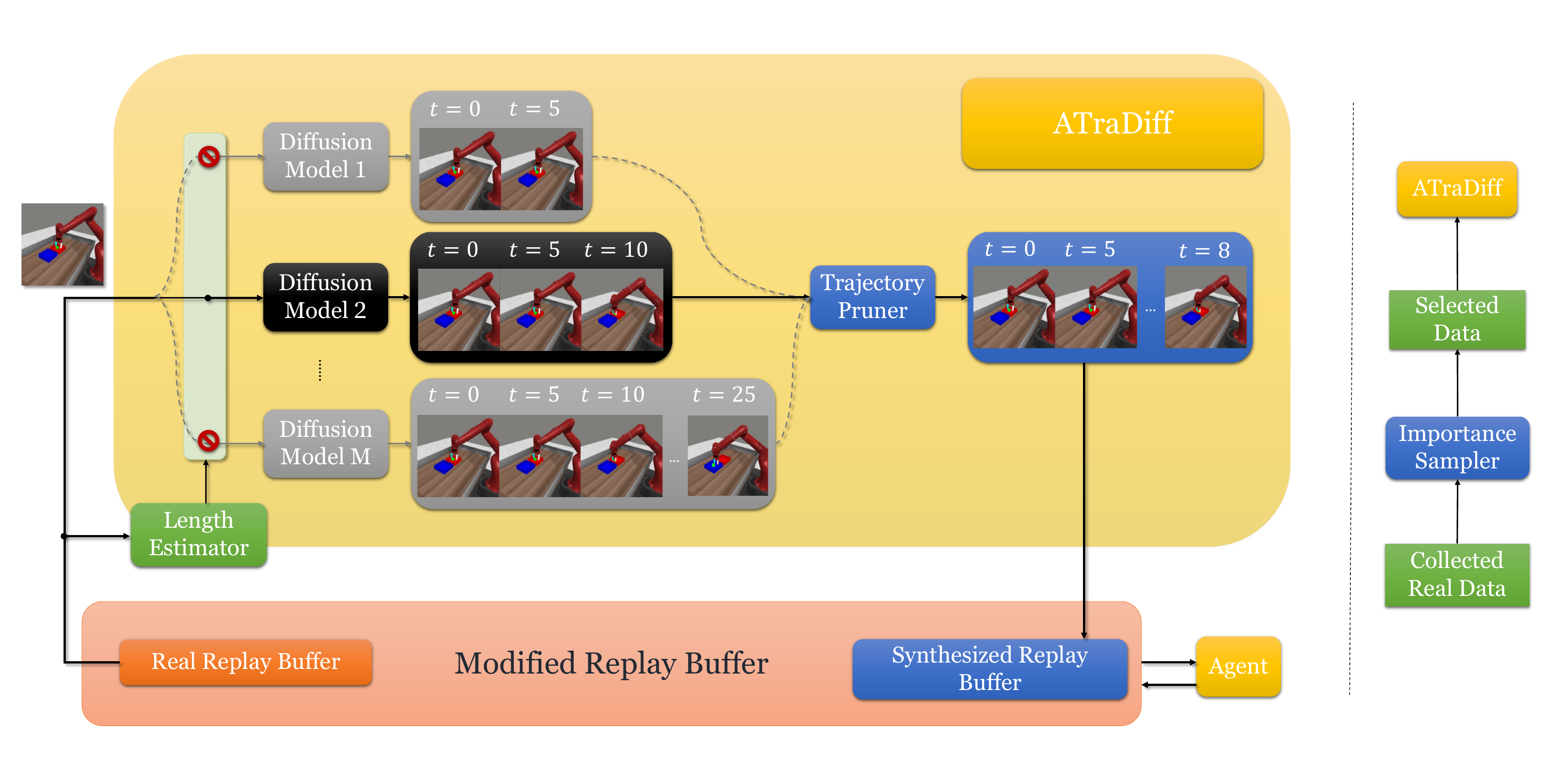}
    \vspace{-0.1in}
    \caption{Illustrative overview of our ATraDiff framework. \textbf{Left:} A diffuser containing multiple diffusion models, a length estimator, and a trajectory pruner.   \textbf{Right:} Workflow of the online adaptation.}
    \vspace{-.5cm}
    \label{figure:paradigm}
\end{figure*}

We now present our approach to accelerating online reinforcement learning by training our generative model ATraDiff on the offline data to synthesize trajectories. We begin by introducing how we design and train ATraDiff (Section~\ref{sec:train}) and then explain how we apply ATraDiff to accelerate online reinforcement learning (Section~\ref{sec:deploy}). Finally, we introduce how this generator could be dynamically adapted during the online training process (Section~\ref{sec:adapt}). Figure~\ref{figure:paradigm} illustrates the overall framework of our approach.

\subsection{Adaptive Trajectory Diffuser}
\label{sec:train}

Our primary objective with the generator is to train a diffusion model $p(\cdot)$  that captures the trajectory data distribution present within the offline data. By doing so, we can synthesize new trajectories $(s_1, a_1, r_1, s_2, a_2, r_2, \dots, s_t, a_t, r_t)$ with the learned diffusion model.  Our work introduces a general data generation framework that is \emph{agnostic} to the specific form of generation: we investigate both \emph{state-level} and \emph{image-level} generation. The state-level generation is to directly generate the states, actions, and rewards from the diffusion model which is trained from scratch. In contrast, image-level generation allows us to benefit from powerful pretrained models, which potentially leads to improved performance. To harness these models more effectively, we advocate for the synthesis of trajectory \textit{images} rather than direct generation of states. Specifically, our strategy entails initially generating images $I_1, I_2, I_3, \dots, I_k$ spanning $k$ continuous frames, where $k$ is a fixed generation length preset for a given generator. Subsequently, we can get the state, action, and reward $s_i, a_i, r_i$ from the resultant images using a trained \textit{decoder} that converts images into states. Conversely, to provide images for our generator, we use another \textit{encoder} to convert states into images. We implement the \textit{encoder} by simple ray tracing~\citep{DBLP:books/daglib/0002788}. More details are provided in Appendix~\ref{appendix:alg}.

\textbf{Image diffusion models.} For the architecture design of our image diffuser, we use Stable Diffusion~\citep{rombach2022highresolution} as the pretrained model and fine-tune it on the specific dataset. To generate images of $k$ continuous frames, we concatenate the $k$ single images $I_1, I_2, \dots, I_k$ into one larger 2D image $\mathcal{I}$ by aligning them in order and use the diffusion model to generate the concatenated image $\mathcal{I}$. Furthermore, we use the information of the first frame as the generation condition, including the current state, task, etc., so that the generated trajectory will be closely related to our learning process, where online learning could benefit more from these synthesized data. 

% \vspace{0.15cm}
\textbf{End-to-end decoder.} 
We design a decoder to decode states, actions, and rewards from the generated images. 
The decoder is a neural network consisting of two components. The first component incorporates the feature distilled from the diffusion layer~\citep{DBLP:conf/iccv/ZhaoRLLZL23}. In the second component, separate multilayer perception~(MLP) networks are used to determine the 3D positions of specific key points, which are used to generate the real states. This two-component design of the decoder allows for efficient processing of the images, by first identifying relevant features through the diffusion model and then pinpointing key points' positions using MLP networks, so that we can calculate the proprioceptive states. The rewards and actions are predicted from the features and the predicted states with separate MLP networks.

\textbf{Generation of Trajectories with Flexible Lengths.} Due to the property of diffusion models, our diffuser described above can only obtain trajectories with a fixed length of $k$; however, to generate a complete trajectory where the horizon is indefinite, we would require the diffuser to output trajectories with flexible lengths, which becomes the crucial problem. To solve this, we introduce a simple \emph{coarse-to-precise} strategy.
We initially train multiple ($M$) diffusion models with varying lengths, such as $k=5, 10, 15, \dots$. 
Before generation, we 
first estimate the required length for the current state with a pretrained network and round it up to the nearest preset length, minimizing the occurrence of redundant transitions. To prune the redundant segment after generation, we introduce a \emph{non-parametric} algorithm to identify the best ending position. 

More concretely, for a generated trajectory $(s_1, s_2, \dots, s_k)$, we first calculate the similarity between each of two adjacent states, $\text{sim}(s_1, s_2), \text{sim}(s_2, s_3), \dots, \text{sim}(s_{k-1}, s_k)$. Then we compute the prefix average $\text{pre}_i$ and suffix average $\text{suf}_i$ of this similarity sequence, where $\text{pre}_i = \frac{\sum_{j=1}^i \text{sim}(s_j, s_{j+1})}{i}, \text{suf}_i = \frac{\sum_{j=i}^k \text{sim}(s_{j-1}, s_j)}{k-i+1}$. Then we get the difference between the prefix average and suffix average $|\text{pre}_i - \text{suf}_i|$ of each position $i$ ($1< i< k$), and find the one with the largest difference to be the ending position. 

This trajectory pruning algorithm operates under the assumption that the average similarity before the ending point should be significantly lower than that after the ending point. This is achieved by training the diffusion model in a way that produces nearly duplicated frames within a redundant segment. To this end, we preprocess the training dataset to \emph{explicitly} introduce redundant segments. Specifically, when training the diffusion model with a generation length of $k$, given a full trajectory $(I_1, I_2, \dots, I_t)$, we first append $k$ additional frames identical to the last frame $I_t$ after the end of the trajectory, resulting in $(I_1, I_2, \dots, I_t, I_t, I_t, \dots, I_t)$. Subsequently, we sample $t$ different sub-trajectories from the padded trajectory, each with a length of $k$, where the starting position of the sub-trajectory ranges from $1$ to $t$. Hence, for the sub-trajectories starting after time-step $t-k$, their final several frames would always be the same.

\subsection{Diffuser Deployment}
\label{sec:deploy}
With the diffuser of ATraDiff explained, we now delve into how the diffuser can be seamlessly integrated with any online RL method with a replay buffer. Intuitively, ATraDiff augments the replay buffer with the data synthesized by its diffuser, and leaves the RL algorithm itself untouched; thus, our approach is orthogonal to any online RL method equipped with a replay buffer.

\begin{algorithm}[t]
\caption{Modified Replay Buffer for RL}

\begin{algorithmic}[1]
\REQUIRE $D= (D_s, D_o, \rho, L)$, ATraDiff.

\FUNCTION{Store $D$, $z = (s, a', s', r)$}

    \STATE ReplayBufferStore($D_o$, $z$)
    %\State Store the transition $z$ into the replay buffer $D_o$.
    \IF{with probability $\frac{\rho}{(1-\rho)L}$}
        \STATE $(s_1, a_1, r_1, \dots, s_l, a_l, r_l) \gets $ ATraDiff($s'$)
        % \State Synthesize a new trajectory $(s_1, a_1, s_2, a_2, \dots, s_t, a_t)$ with ATraDiff, conditioned on state $s'$.
        \FOR{$\forall i$}
            \STATE $z_i = (s_i, a_i, s_{i+1}, r_i)$
            \STATE ReplayBufferStore($D_s$, $z_i$)
        \ENDFOR
        
    \ENDIF
\ENDFUNCTION

\STATE
\FUNCTION{Sample $D$}
    \IF{with probability $\rho$}
        \STATE $z \gets$ ReplayBufferSample($D_s$)
    \ELSE
        \STATE $z \gets$ ReplayBufferSample($D_o$)
    \ENDIF
    \STATE return $z$
\ENDFUNCTION

\end{algorithmic}
\label{alg:rep}
\end{algorithm}

More specifically, consider any RL algorithm with a replay buffer denoted as $D_o$. Typically, RL methods engage with the replay buffer through two primary actions: \textit{store}, which archives a new transition into the replay buffer, and \textit{sample}, which extracts a transition randomly from the replay buffer. With this in mind, we substitute the original buffer $D_o$ with a new replay buffer $D=D_o\cup D_s$, where $D_s$ is the augmenting buffer synthesized by ATraDiff. This modified replay buffer $D$ is characterized by two hyperparameters: $\rho\in[0, 1]$, denoting the probability of sampling from synthesized data $D_s$ in RL, and $L\in\mathbb{N}$, indicating the expected length of synthesized trajectories. 

Whenever we \textit{store} a transition $(s, a, s', r)$ into $D$ during the RL algorithm, the following three steps are performed:
    \vspace{-0.15cm}
\begin{itemize}
    \item We first \textit{store} it into $D_o$;
    \item With probability $\frac{\rho}{(1-\rho)L}$, a full trajectory $(s_1, a_1, r_1 s_2, a_2, r_2 \dots, s_l, a_l, r_l)$ is synthesized with the diffuser, with state $s'$ as the initial state (i.e., $s_1 = s'$). Note the probability $\frac{\rho}{(1-\rho)L}$ is designed to keep the ratio between the total size of $D_s$ and $D_o$ to be $\frac{\rho}{1-\rho}$; 
    \item All synthesized transitions $(s_i, a_i, s_{i+1}, r_i)$ will be \textit{stored} to the replay buffer $D_s$.
\end{itemize} 

    \vspace{-0.15cm}
When we \textit{sample} from $D$, we \textit{sample} from $D_s$ with probability $\rho$ and from $D_o$ with probability $1-\rho$. See Alg.~\ref{alg:rep} for pseudo-code. Note that the sampling process can be arbitrary, i.e., our method is also orthogonal to other sampling techniques such as prioritized buffer~\citep{schaul2015prioritized}. 

In our algorithm, the expected length, denoted as $L$, and the actual length, denoted as $l$, serve distinct purposes and are derived through different methods. The expected length $L$ is defined as the mean length of all trajectories generated during the training phase. It is empirically determined by averaging the lengths of all trajectories generated in preliminary empirical experiments, conducted before the actual training phase. As a critical hypeparameter, $L$ guides the total number of transitions generated by ATraDiff, ensuring a controlled and predictable size of the generated dataset. Conversely, the actual length $l$ is determined subsequent to the generation process. Within this phase, the appropriate generator is selected via the length estimator, followed by trajectory pruning to achieve the desired length. The value of $l$ is inherently variable, adjusting across different generation instances to align with the dynamic contexts and specific requirements of each. Over the course of the training phase, we expect that the mean of all actual lengths $l$ to converge towards the predefined expected length $L$.

\subsection{Online Adaptation}
\label{sec:adapt}

Although the fixed ATraDiff can improve the performance of RL methods in some simple environments, we may still face the problem of distribution shift between the evaluation task and the offline data in complicated environments. To overcome this issue and further improve the generation quality, we propose an online adaptation technique by continually training the diffusion model on new experiences. 

Concretely, ATraDiff is periodically updated on the real transitions stored in $D_o$ and then used to generate new trajectories. Meanwhile, we keep a copy of the original version of the diffuser to mitigate potential catastrophic forgetting. 

Furthermore, we use more valuable samples to adapt our ATraDiff during online training. Specifically, we design an \emph{indicator} to measure the importance of each sample for our online learning, and a \emph{pick-up} strategy to choose samples from $D_o$. By default, we are introducing two indicators: the TD-error indicator, and the Reward indicator. For the TD-error indicator, the importance of a transition $(s, a, s', r)$ is defined to be $|r + \gamma \max_{a'}Q(s', a') - Q(s, a)|$, where $Q$ is the learned value function. The TD-error indicator performs better in most cases, despite that it could only be used in some value-based RL methods. The Reward indicator would be more general to all RL methods, as the importance of a transition is defined to be the total reward collected in the full trajectory. The primary pick-up strategy is to maintain a subset of samples with higher importance, but part of the samples are always used to update the diffuser. Hence, we randomly drop some samples from the maintained subset regardless of their importance. We further conduct experiments to analyze the effectiveness of our indicator and pick-up strategy and the impact of different design choices in the ablation study. 

\vspace{.5mm}

\section{Experiments}

We conduct comprehensive experiments to evaluate the effectiveness of our data generator ATraDiff. First, we validate that our approach is able to improve the performance of both basic and state-of-the-art online RL methods by combining them with our ATraDiff (Section~\ref{sec:exp-online}). Next, we show that our method can further improve the performance of a variety of state-of-the-art offline-to-online RL algorithms in complicated environments using online adaptation (Section~\ref{sec:exp-off2on}). In addition, we evaluate our method in the offline setting to show that ATraDiff achieves promising results in offline data augmentation as well (Section~\ref{sec:exp-off}). Finally, we conduct some ablation studies to validate the effectiveness of different components in our approach (Section~\ref{sec:ablation}). For evaluation, all results in this section are presented as the median performance over 5 random seeds along with the 25\%-75\% percentiles.

\subsection{ATraDiff Improves Online RL}
\label{sec:exp-online}
We first show that the performance of state-of-the-art online RL methods can be improved by our ATraDiff learned with the offline data. We consider 3 environments from D4RL Locomotion~\citep{fu2020d4rl}, including 12 different offline data with varying levels of expertise. For comparison, we choose SAC~\citep{sac} as the basic online RL algorithm and REDQ~\citep{redq} as the state-of-the-art sample-efficient algorithm. We run both baselines and their variants combined with our ATraDiff for 250K steps. 

\begin{figure}[t]
    \centering
    \includegraphics[width=\linewidth]{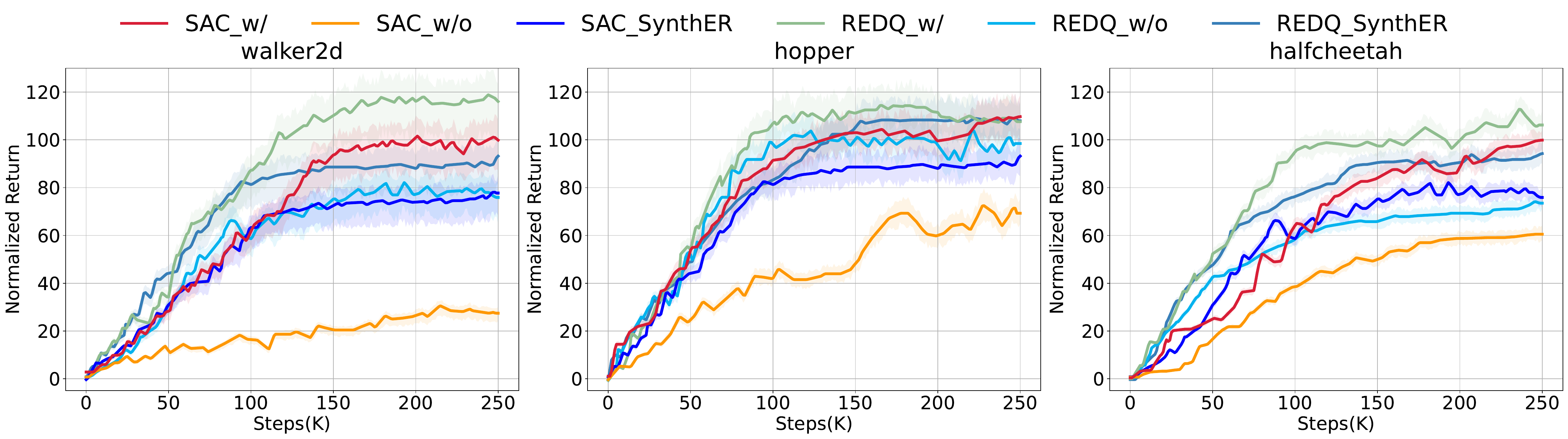}
    \vspace{-2em}
    \caption{Learning curves of online RL on the D4RL Locomotion benchmark. ATraDiff (denoted as `w/') consistently and significantly improves the performance of the two representative RL methods across all three environments, irrespective of whether basic or advanced algorithms are employed. ATraDiff also outperforms SynthER which synthesizes transitions. These results validate the effectiveness and generalizability of our diffuser.}
    \label{figure:offline}    
\end{figure}

The overall result is summarized in Figure~\ref{figure:offline}, showcasing the average performance of each environment with 4 different offline datasets. Detailed performance breakdowns are provided in Appendix~\ref{appendix:full_result}. We see that the performance of the two baselines is both significantly improved by our ATraDiff, especially on the halfcheetah and walker2d environments. This validates the strength of ATraDiff. If we run these online RL methods for enough time, they can also achieve comparable results, while ATraDiff improves the sample efficiency. Here, we only utilize the fixed diffuser instead of using online adaptation, which indicates that our fixed diffuser can already accelerate online RL in some simple environments.

\subsection{ATraDiff Improves Offline-to-Online RL in Complicated Environments}
\label{sec:exp-off2on}
In this section, we show that ATraDiff with online adaptation can improve the performance of the state-of-the-art offline-to-online RL methods. We further consider the following environments:

\textbf{D4RL AntMaze}~\citep{fu2020d4rl}\textbf{.} There are 6 sparse reward tasks that require the agent to learn to walk with controlling an 8-DoF Ant robot and navigate through a maze to reach the goal. 
    
\textbf{D4RL Kitchen}~\citep{fu2020d4rl}\textbf{.} A simulated kitchen environment is proposed by \citet{gupta2019relay}, which involves controlling a 9-DoF robot that manipulates different objects in a kitchen environment (e.g., slide cabinet door, switch overhead light, and open microwave). The downstream task is to complete a sequence of multiple subtasks in order with a sparse, binary reward for each successfully completed subtask. The offline dataset only contains part of the full sequence, meaning that the agent needs to learn to compose sub-trajectories.
    
\textbf{Meta-World}~\citep{DBLP:conf/corl/YuQHJHFL19}\textbf{.} By combining the modified tasks using a single camera viewpoint consistently over all the 15 subtasks generated by~\citet{DBLP:conf/corl/SeoHLLJLA22}, we create two challenging tasks. The first one is a multi-task setting, where the offline data contain the trajectories of 14 subtasks and 1 different evaluation subtask. The second one is a harder version of the D4RL Kitchen task, where the agent needs to complete a sequence of 8 subtasks with medium difficulty in the correct order, while the offline data only contain trajectories for single subtasks. The environment will be set to the initial state of the next subtask, when the previous one was completed by the agent. This design ensures that the agent progresses through the tasks in the predetermined order, with objects from different tasks kept isolated to prevent interference. For both tasks, the trajectories of any single subtask consist of the provided data and collected data from the online training.

We consider a set of strong baselines from prior work on offline-to-online RL, including the skill-based method (SPiRL~\citep{pertsch2020accelerating}), behavior-prior-based method PARROT~\citep{kumar2022datadriven}, and balanced sampling method RLPD~\citep{ball2023efficient}. We run these baselines and their variants combined with our diffuser. To show the effectiveness of our method, we also compare it with SynthER~\citep{lu2023synthetic}. For the following results, the curve named ``SynthER'' shows the best performance of SynthER combined with any of the baselines.

\begin{figure}[t]
    \centering
    \includegraphics[width=\linewidth]{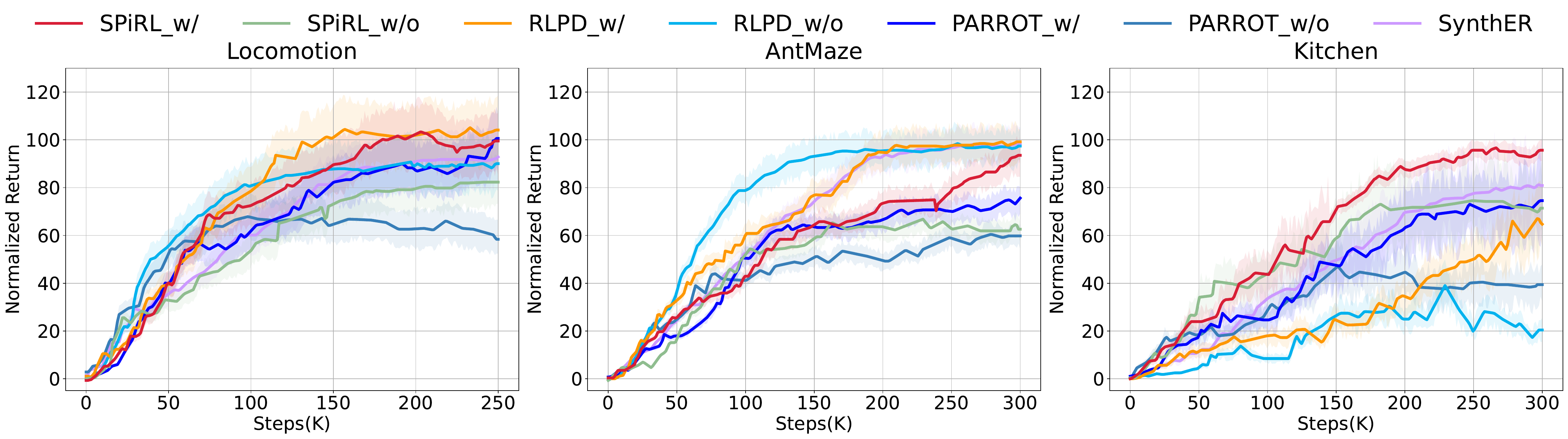}
        \vspace{-2em}
    \caption{Learning curves of offline-to-online RL on the D4RL benchmark. ATraDiff (denoted as `w/') further boosts the performance of advanced and recent offline-to-online RL baselines across all three environments, leading to state-of-the-art results especially in complex settings, where the improvements are particularly noteworthy. This shows the importance of our online adapted diffuser. The curve named ``SynthER'' shows the best performance of SynthER combined with any of the baselines.}
    \label{figure:off2on}
    % \vskip{-2em}
\end{figure}

\begin{figure}[h]
    \centering
            \vspace{-1em}
    \includegraphics[width=\linewidth]{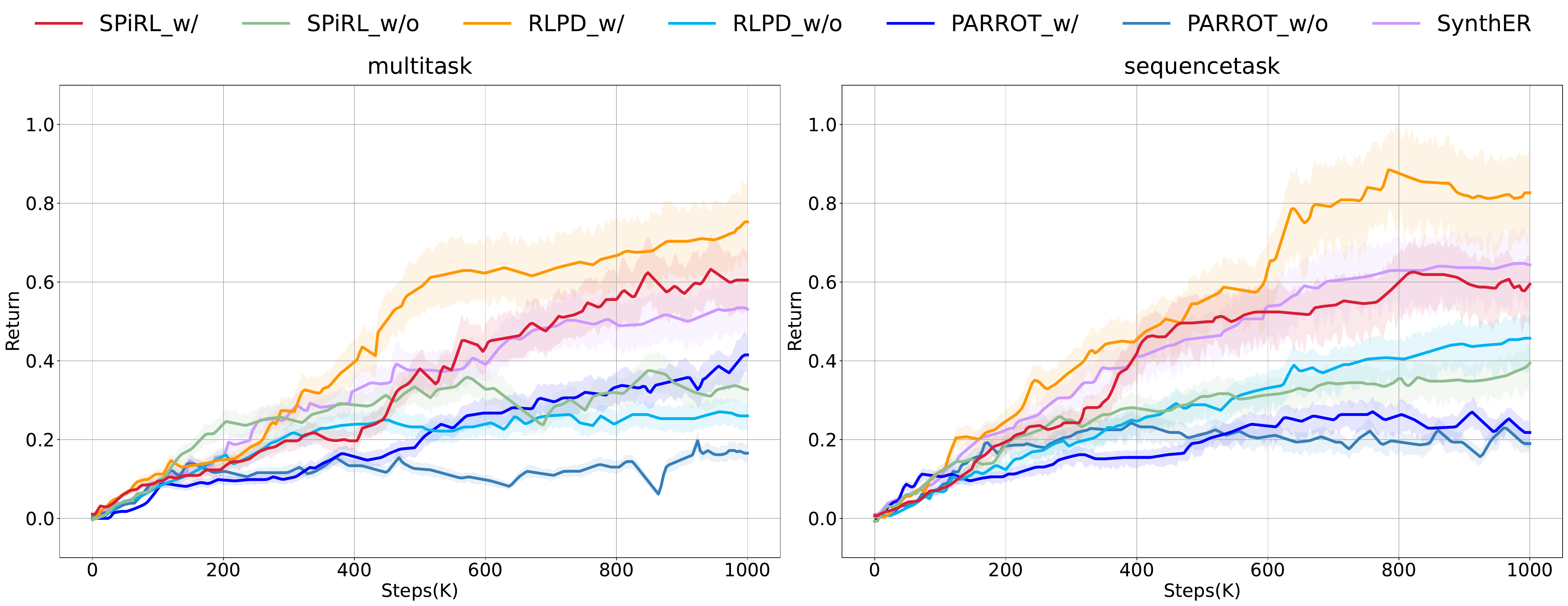}
        \vspace{-2em}
    \caption{Learning curves of offline-to-online on the Meta-World benchmark. While the two tasks within the Meta-World environment are designed purposefully to be very changeling with considerable distribution shifts, ATraDiff (denoted as `w/') is still effective and significantly improves the performance of advanced and recent offline-to-online RL baselines. This further validates the strength of ATraDiff in tacking distribution shifts between offline data and online tasks. The curve named ``SynthER'' shows the best performance of SynthER combined with any of the baselines.}
    \label{figure:aba_meta}
\end{figure}

As shown in Figure~\ref{figure:off2on}, our diffuser significantly improves the performance of some methods, and obtains at least comparable results for all methods. For all these environments, the best result is always achieved by some method combined with our diffuser, which shows that our online adapted diffuser can further improve the performance of current offline-to-online methods. Notably, we introduce two advanced tasks within the Meta-World environment to emphasize the challenges posed by distribution shifts. Our motivation is to understand how our method responds under conditions where distribution shift problems are more pronounced. As demonstrated in Figure~\ref{figure:aba_meta}, our approach exhibits superior performance, suggesting that our method holds significant promise in effectively addressing such shifts.
% Full results please refer to Appendix~\ref{appendix:full_result}.

\subsection{ATraDiff Improves Offline RL}
\label{sec:exp-off}

Although ATraDiff is primarily intended for offline-to-online scenarios, it also demonstrates strong performance in offline data augmentation. This section details experiments conducted to assess ATraDiff's effectiveness in an exclusively offline environment.

Our objective is to verify whether ATraDiff can match or surpass the results achieved in previous research, specifically referencing the work SynthER~\citep{lu2023synthetic} which synthesizes \textit{transitions}. To ensure a fair and accurate comparison, we meticulously replicated the experimental settings used in \citet{lu2023synthetic}. We also compare with the previous state-of-the-art augmentation method S4RL~\citep{s4rl}. Our testing ground is the D4RL Locomotion benchmark~\citep{fu2020d4rl}. We extend the original dataset to $5$M samples, following the settings in \citet{lu2023synthetic}.
%as detailed in 

The findings, as outlined in Table~\ref{table:offline}, are quite revealing. They indicate that our ATraDiff method generally outperforms other baselines across various datasets. This performance improvement is particularly pronounced in high-quality datasets, underscoring the effectiveness of our approach in generating helpful trajectories and the benefits of \textit{trajectory-level generation over transition-level generation}.

\begin{table*}[h!]\tiny
    \centering
    \resizebox{.9\textwidth}{!}{
    \begin{tabular}{c|c|c|c|c|c|c|c|c}
    \hline
    \textbf{Task Name} & \textbf{TD3+BC} & \parbox[t][.5cm]{.8cm}{\textbf{TD3+BC\\+SynthER}} & \parbox[t][.5cm]{.7cm}{\textbf{TD3+BC\\+S4RL}} & \parbox[t][.5cm]{.7cm}{\textbf{TD3+BC\\+ATraDiff}} & \textbf{IQL} & \parbox[t][.5cm]{.7cm}{\textbf{IQL+\\SynthER}} & \parbox[t][.5cm]{.6cm}{\textbf{IQL+\\S4RL}} & \parbox[t][.5cm]{.7cm}{\textbf{IQL+\\ATraDiff}}  \\ \hline
    halfcheetah-random & 11.3 & 12.2 & 11.5 & 12.5 & 15.2 & \textbf{17.2} & 15.8 & 17.1 \\
    halfcheetah-medium & 48.1 & 49.9 & 48.5 & 52.3 & 48.3 & 49.6 & 48.8 & \textbf{53.1}  \\ 
    halfcheetah-replay & 44.8 & 45.9 & 45.9 & 46.5 & 43.5 & 46.7 & 46.3 & \textbf{49.2}  \\ 
    halfcheetah-expert & 90.8 & 87.2 & 91.2 & 93.6 & 94.6 & 93.3 & 94.3 & \textbf{95.2}  \\ \hline
    hopper-random & 8.6 & 14.6 & 9.4 & \textbf{15.2} & 7.2 & 7.7 & 7.4 & 8.1 \\
    hopper-medium & 60.4 & 62.5 & 63.4 & 65.7 & 62.8 & 72.0 & 70.3 & \textbf{72.4}  \\ 
    hopper-replay & 64.4 & 63.4 & 62.3 & 64.7 & 84.6 & 103.2 & 95.6 & \textbf{103.6}  \\ 
    hopper-expert & 101.1 & 105.4 & 103.5 & 111.2 & 106.2 & 110.8 & 108.1 & \textbf{113.6}  \\ \hline
    walker-random & 0.6 & 2.3 & 3.2 & 2.1 & 4.1 & 4.2 & 4.1 & \textbf{4.3} \\
    walker-medium & 82.7 & 84.8 & 83.7 & 87.5 & 84.0 & 86.7 & 84.5 & \textbf{89.1}  \\ 
    walker-replay & 85.6 & \textbf{90.5} & 88.3 & 86.3 & 82.6 & 83.3 & 83.1 & 85.4  \\ 
    walker-expert & 110.0 & 110.2 & 106.3 & 111.2 &\textbf{111.7} &111.4 & 111.3 & \textbf{111.7}  \\
    \hline
    \end{tabular}
    }
\caption{Normalized return of offline RL on the D4RL Locomotion benchmark. Our ATraDiff outperforms SynthER and S4RL on almost all the tasks and datasets. While primarily intended for offline-to-online scenarios, ATraDiff also demonstrates strong performance
in offline data augmentation.}
\vspace{-.2cm}
\label{table:offline}
\end{table*}

\subsection{Ablation Study and Analysis}
\label{sec:ablation}
In this section, we carry out ablation studies to justify the effect of different components of our ATraDiff. Additional ablation studies are provided in Appendix~\ref{appendix:atra_syn} and the generated trajectories are visualized in Appendix~\ref{appendix:vis}.

\begin{figure}[h]

    \centering
    \includegraphics[width=.95\linewidth]{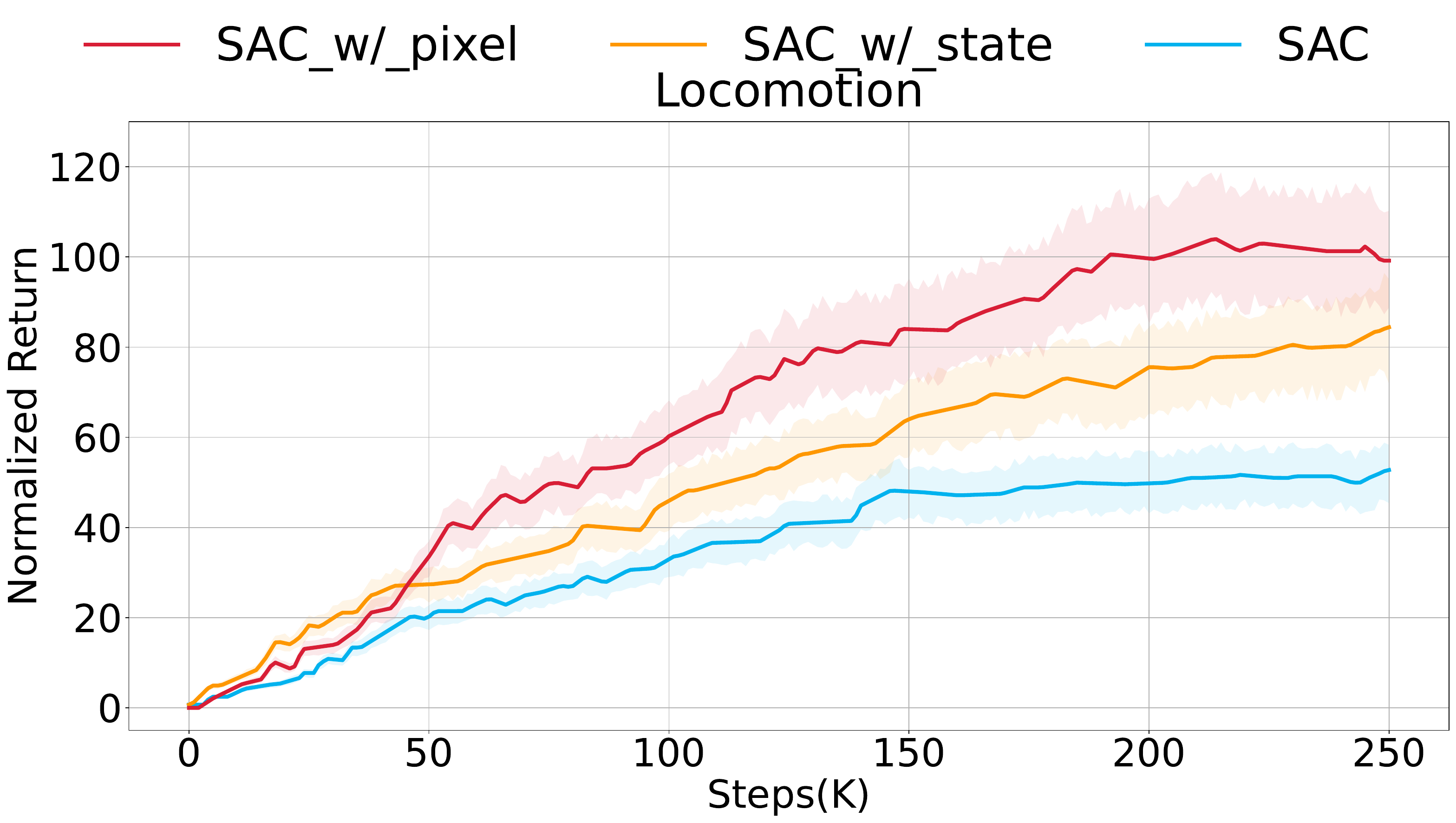}
    \caption{Ablation study on image-level generation and state-level generation. The image-level diffuser outperforms the state-level diffuser in complicated tasks, with noticeable performance gains.}
    \label{figure:aba_image}
    \vspace{-.8cm}
\end{figure}

\textbf{Image-level generation vs. state-level generation.} Our work introduces a general data generation method that is agnostic to the specific form of generation. As shown in Figure~\ref{figure:aba_image}, both image and state generation variants significantly outperform the baselines, while the image-level diffuser generally outperforms the state-level diffuser on the D4RL Locomotion environment. We usually need to synthesize longer and more flexible trajectories in such complicated environments, where the image-level diffuser performs better compared with simpler diffusion models. Our hypothesis is that image generation can benefit from a pretrained generation model (e.g., Stable Diffusion) while state generation is trained from scratch, though image generation needs additional modules to convert between images and states.

Meanwhile, we conduct an analysis to demonstrate the effectiveness and generalizability of our ATraDiff on pixel-based reinforcement learning methods. To this end, we employ ATraDiff to upsample the dataset from the V-D4RL benchmark~\citep{vd4rl}, and use Behavior Cloning (BC)~\citep{BC} as the baseline to test on the benchmark. The results shown in Table~\ref{tab:vrl} validate that our algorithm is also applicable to the visual reinforcement learning methods, consistently improving their performance.

\begin{table}[H]
    \centering
    \scriptsize
    \renewcommand{\arraystretch}{0.9} % Adjust row height
    \resizebox{.485\textwidth}{!}{\begin{tabular}{c|*{4}{>{\centering\arraybackslash}p{0.4cm}}|*{4}{>{\centering\arraybackslash}p{0.4cm}}}
        \toprule
        \textbf{Task Name} & \multicolumn{4}{c|}{walker} & \multicolumn{4}{c}{cheetah} \\
        \midrule
        \textbf{Dataset} & mixed & medium & medexp & expert & mixed & medium & medexp & expert \\
        \midrule
        \textbf{BC} & 16.5 & 40.9 & 47.7 & 91.5 & 25.0 & 51.6 & 57.5 & 67.4 \\
        \midrule
        \textbf{BC+ATraDiff} & 17.7 & 42.2 & 50.2 & 93.1 & 25.2 & 55.3 & 68.1 & 85.3 \\
        \bottomrule
    \end{tabular}
}    \caption{Normalized return of offline RL on the V-D4RL benchmark. Our ATraDiff consistently improves the performance of pixel-based reinforcement learning methods.}
    \label{tab:vrl}
    \vspace{-0cm}
\end{table}

\begin{figure}[h]
    
    \centering
    \includegraphics[width=\linewidth]{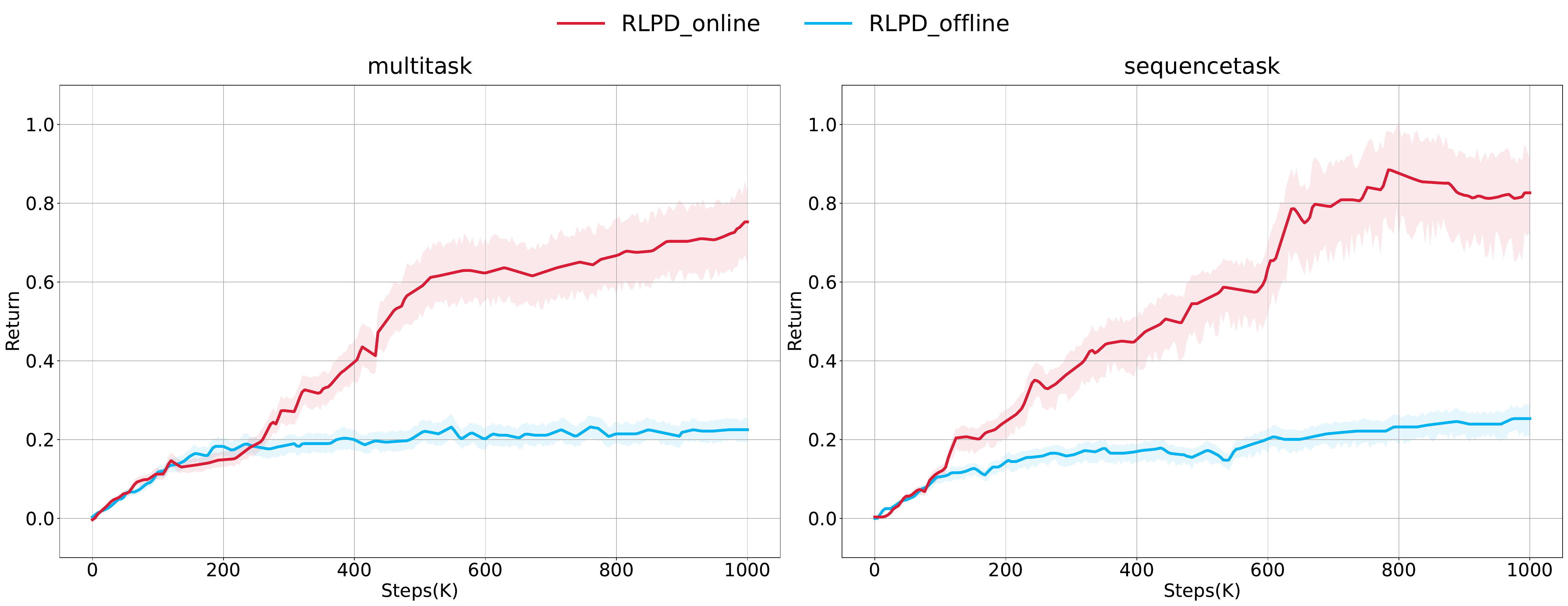}
    \vspace{-2em}
    \caption{Ablation study on online adaptation of ATraDiff. In simple tasks, we observe that the online diffuser achieves results comparable to the fixed diffuser. However, online adaptation mitigates the problem of data distribution shift and becomes critical in complicated environments.}
    \label{figure:aba_online}
\end{figure}

\textbf{Is online adaptation beneficial for our diffuser?} We now examine the effect of online adaptation on performance. We revisit the experiments in Section~\ref{sec:exp-off2on} and replace the online adapted diffuser with a fixed one. As shown in Figure~\ref{figure:aba_online}, we find that the online diffuser significantly outperforms the fixed diffuser in complicated tasks. In tasks of D4RL Locomotion and D4RL AntMaze, the online diffuser achieves comparable results to the fixed diffuser. However, in the task of D4RL Kitchen and two tasks in Meta-World, the online diffuser has fully demonstrated its superiority, which validates that the online adaptation can indeed mitigate the problem of data distribution shift and thus focus on the evaluation task.

\begin{figure}[h]
    
    \centering
        \vspace{-1em}
    \includegraphics[width=\linewidth]{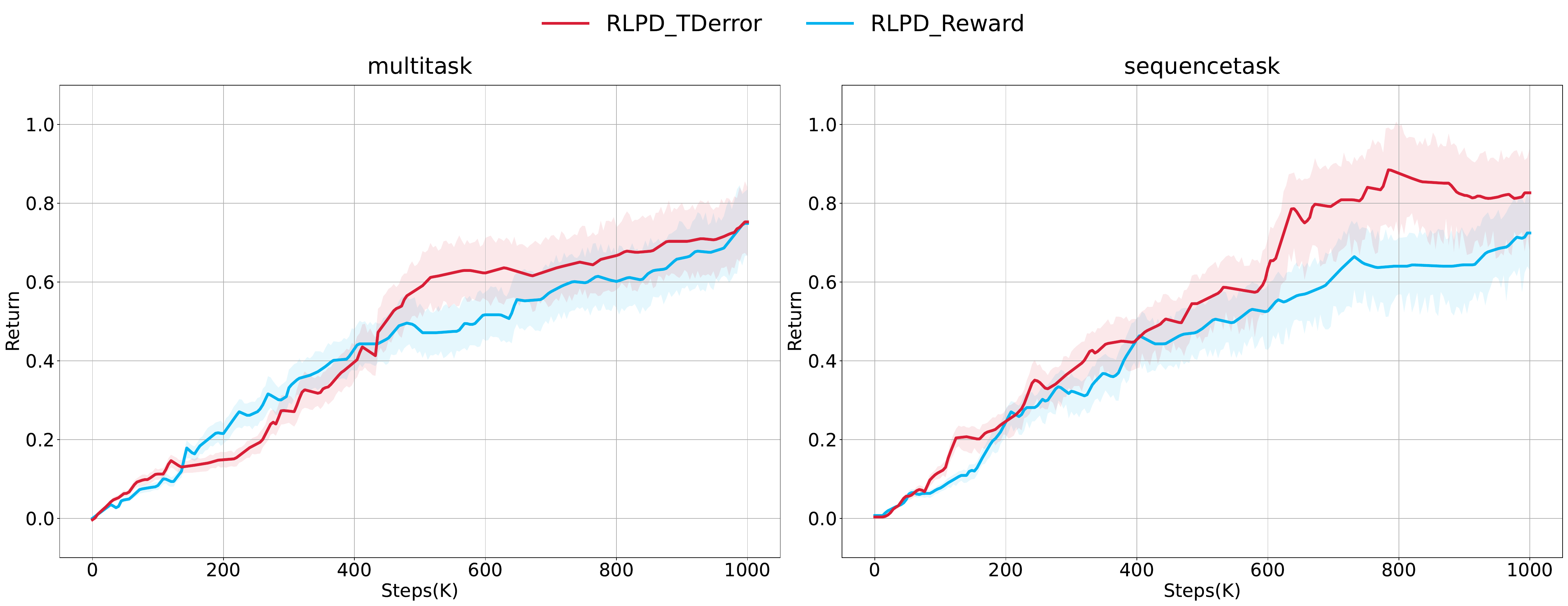}
    \vspace{-2em}
    \caption{Ablation study on the importance indicator together with its associated pick-up strategy. Different types of indicators lead to varied behavior and performance. While the Reward indicator is in principle more general to RL methods, the indicator based on TD-error achieves better performance in the complicated Meta-World benchmark.}
    \label{figure:aba_ind}
\end{figure}

%\vspace{-.5cm}
\textbf{Effect of different importance indicators over online adaptation.} We show that different online adaptation strategies have noticeable impact on performance. Our experiments focus on the indicator used to measure the importance of collected samples together with its associated pick-up strategy for selecting samples according to this importance. We include the Reward indicator and its corresponding pick-up strategy. The result shown in Figure~\ref{figure:aba_ind} indicates that the TD-error indicator outperforms the Reward indicator.

\begin{figure}[h]
    
    \centering
    \includegraphics[width=\linewidth]{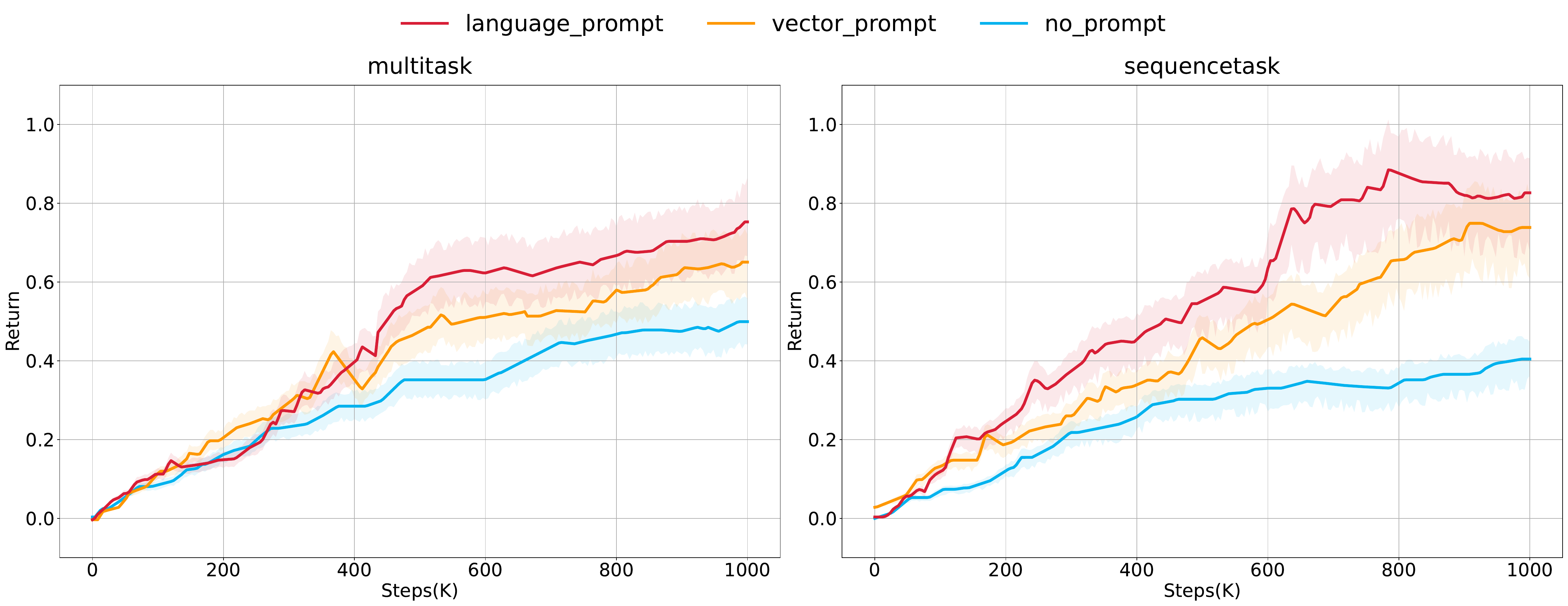}
    \vspace{-.8cm}
    \caption{Ablation study on design choices of task prompts. Different types of task prompts lead to varied behavior and performance. }
    \label{figure:aba_prompt}   
    %\vspace{-.4cm}
\end{figure}

\textbf{Effect of different task prompts.} We show that different task prompts noticeably affect performance. To evaluate this, we conduct experiments on Meta-World, testing three strategies: language task prompt, one-hot task prompt, and no prompt, where the one-hot prompt simply uses a one-hot vector to represent different tasks. The result shown in Figure~\ref{figure:aba_prompt} demonstrates that the language prompt outperforms the one-hot prompt, and both significantly exceed the performance of the baselines without any prompt.

\vspace{-0cm}

\begin{figure}[h]
    
    \centering
    \includegraphics[width=.9\linewidth]{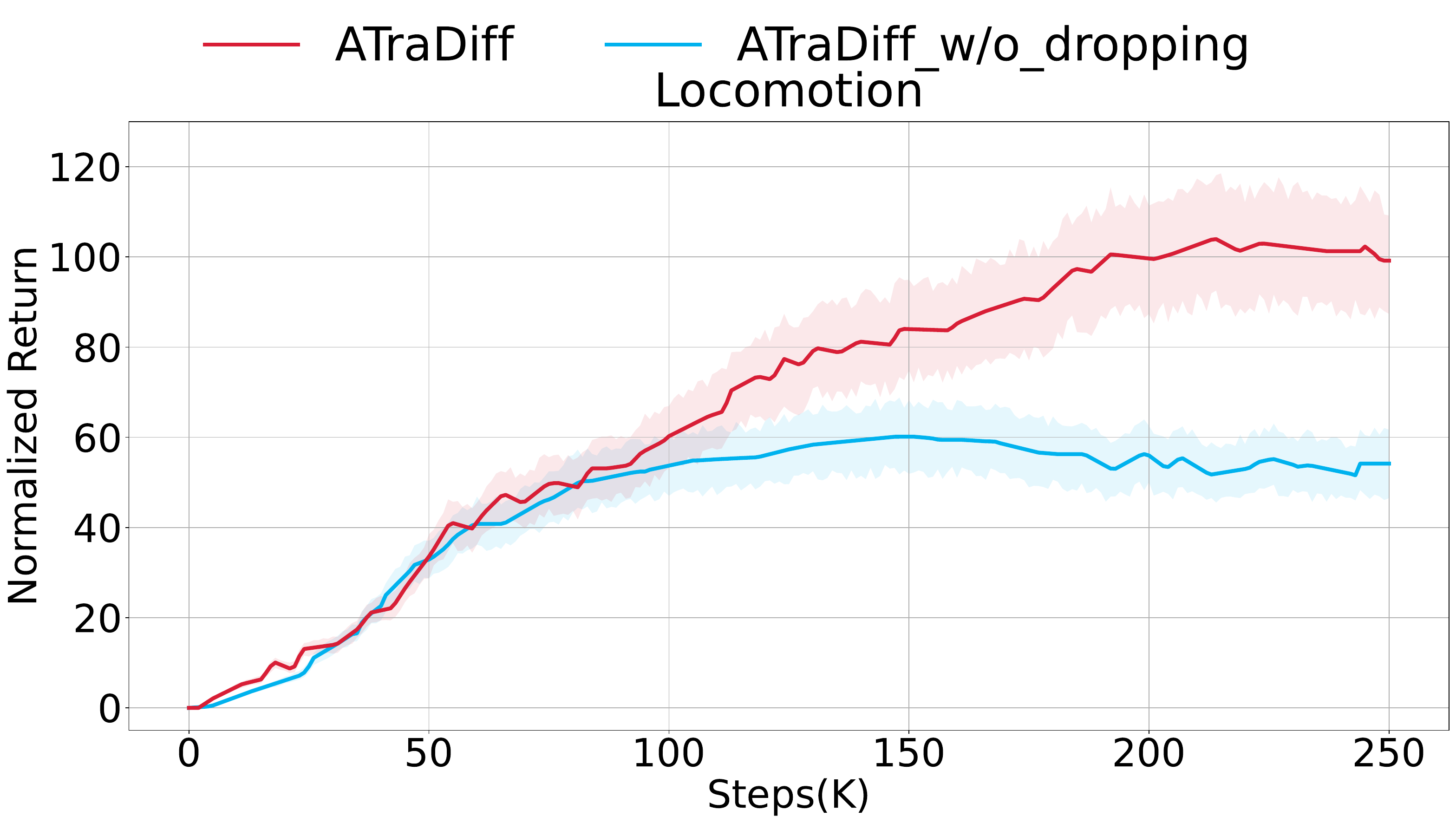}
   % \vspace{-.8cm}
    \caption{Ablation study on the random dropping strategy. We observe that the random dropping strategy can significantly improve the performance, which validates that different strategies indeed influence the performance.}
    \label{figure:aba_drop}   
    \vspace{-.4cm}
\end{figure}

\textbf{Effect of random dropping strategy.} Here we show that the random dropping strategy is very important in the online adaptation process with the Reward indicator. In our online adaptation phase, when using the total reward importance indicator, the importance of any trajectory will never change during the whole training process. Hence, trajectories with high importance might always be used to update the generator, which inspires us to introduce a random dropping strategy. The ablation study conducted on the D4RL Locomotion environment shown in Figure~\ref{figure:aba_drop} illustrates that the random dropping strategy can significantly improve the performance.

\vspace{-0cm}

\begin{figure}[!t]
    
    \centering
    \includegraphics[width=\linewidth]{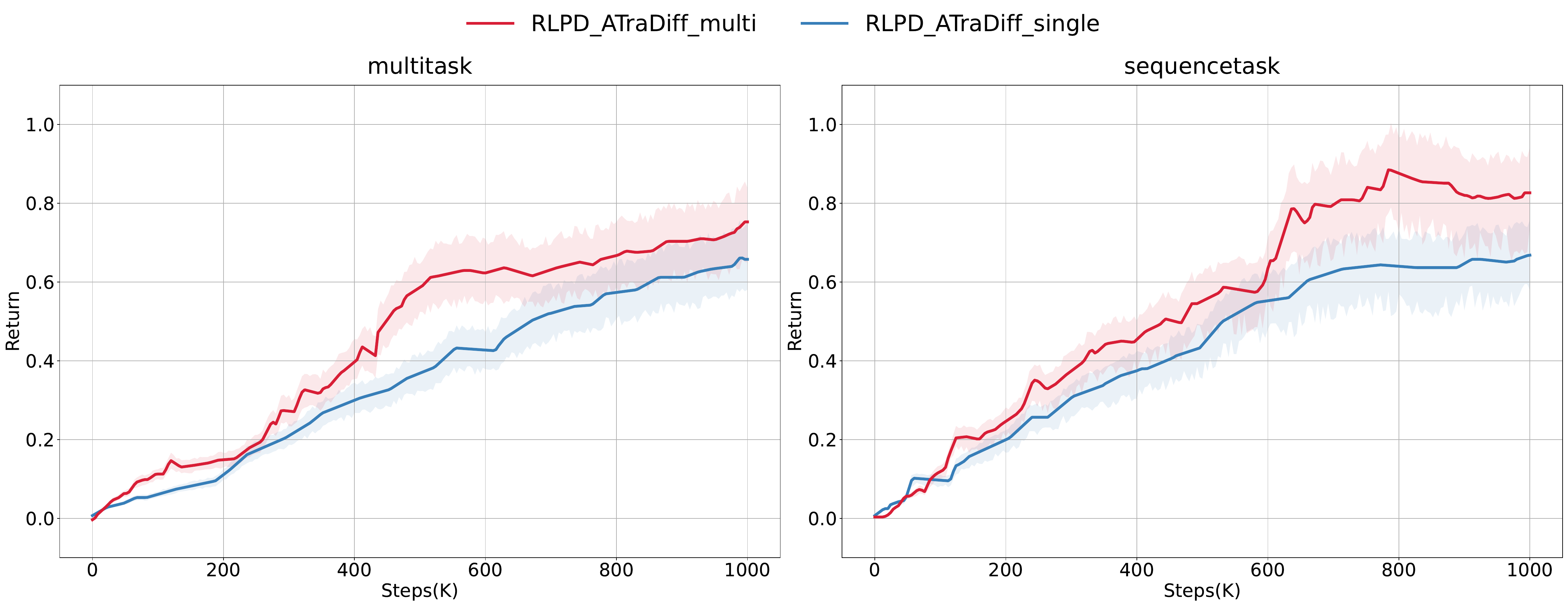}
    \vspace{-.8cm}
    \caption{Ablation study on multiple diffusion models. We observe that the performance of ATraDiff with multiple diffusion models significantly outperforms the fixed single diffusion model.}
    \label{figure:aba_multi}   
    \vspace{-.4cm}
\end{figure}

\textbf{Effect of multi-generator.} Now we show that the multi-generator architecture is crucial. We conduct experiments on Meta-World to test the effect of the multiple diffusion model design. We replace the flexible task-length generation control scheme with a fixed length generator of $25$, which is the largest length in our setting, and directly apply our pruning strategy after the generation. The result shown in Figure~\ref{figure:aba_multi} demonstrates that with multiple diffusion models, the performance of our method significantly outperforms the fixed single diffusion model.

\vspace{-0cm}

\section{Conclusion}

We introduce ATraDiff, a novel diffusion-based approach that synthesizes full trajectories. By training a set of diffusion models that generate trajectories of different lengths and selecting the most suitable model together with a non-parametric pruning algorithm, we obtain a generator that produces trajectories with varied lengths. By applying such a generator to transitions stored in the replay buffer and augmenting the buffer with generated data, ATraDiff can accelerate any online RL algorithm with a replay buffer. In multiple environments, ATraDiff significantly improves existing online, offline, and offline-to-online RL algorithms. We hope that our work is beneficial to the online RL community in addressing the long-standing data efficiency problem.%using imaginary data believe

\textbf{Limitations and future work.} One limitation of ATraDiff is that the training of multiple diffusion models can be computationally intensive (as evaluated in Appendix~\ref{sec:time}), and the problem escalates with longer generated trajectories. Thus, an interesting future direction is to learn a single diffuser that directly generates trajectories of varied lengths instead of selecting from a collection of models.

\section*{Acknowledgements}

This work was supported in part by NSF Grant 2106825, NIFA Award 2020-67021-32799, the Jump ARCHES endowment through the Health Care Engineering Systems Center at Illinois and the OSF Foundation, and the IBM-Illinois Discovery Accelerator Institute. This work used NVIDIA GPUs at NCSA Delta through allocations CIS220014 and CIS230012 from the ACCESS program.

\section*{Impact Statement}

Our research contributes to decision-making processes by leveraging offline data, aiming to enhance the efficiency of reinforcement learning. While our advancements offer significant benefits in terms of task-solving capabilities, they also raise concerns about potential negative societal impacts. For instance, the deployment of highly efficient autonomous systems in the workforce could lead to job displacement and increased economic inequality. Furthermore, our approach relies on a generative model for synthesizing new data, which introduces concerns regarding the authenticity and reliability of the generated outcomes. The possibility of generating fictitious or misleading data underscores the importance of critically assessing and mitigating potential ethical and societal implications, ensuring that advancements in autonomous decision-making contribute positively to society while minimizing adverse effects.

\bibliography{icml2024}
\bibliographystyle{icml2024}

%%%%%%%%%%%%%%%%%%%%%%%%%%%%%%%%%%%%%%%%%%%%%%%%%%%%%%%%%%%%%%%%%%%%%%%%%%%%%%%
%%%%%%%%%%%%%%%%%%%%%%%%%%%%%%%%%%%%%%%%%%%%%%%%%%%%%%%%%%%%%%%%%%%%%%%%%%%%%%%
% APPENDIX
%%%%%%%%%%%%%%%%%%%%%%%%%%%%%%%%%%%%%%%%%%%%%%%%%%%%%%%%%%%%%%%%%%%%%%%%%%%%%%%
%%%%%%%%%%%%%%%%%%%%%%%%%%%%%%%%%%%%%%%%%%%%%%%%%%%%%%%%%%%%%%%%%%%%%%%%%%%%%%%
\newpage
\appendix
\onecolumn
\section{Algorithm Details}
\label{appendix:alg}

\subsection{Pick-up Strategies for Online Adaptation}

In our training methodology, we have designed different pick-up strategies for handling different importance indicators, recognizing their unique properties and implications for the learning process.

For the TD-error indicator, a \textit{dynamic} approach is employed due to the variable nature of its importance, which can change as the critic function updates during training. To manage this, a heap structure is used to maintain a buffer of 50,000 samples. This buffer is continuously updated to reflect the current importance of each sample, ensuring that even those with initially low importance are retained for potential future relevance. When it comes time to update the diffuser, a selection of 5,000 high-importance samples is drawn from this buffer, aligning the update process with the most pertinent data at that moment.

In contrast, the strategy for reward importance takes into account its \textit{static} characteristic -- the sample importance of such an indicator does not change throughout the training. Here, a smaller buffer of 5,000 samples is employed, updated similarly based on importance. However, to counter the risk of high-importance samples perpetually dominating the buffer, a rotation scheme is implemented. After each update, some samples, particularly the older ones, are dropped. This scheme ensures that each sample contributes to the diffuser update for a limited number of times, thus maintaining a fresh and current dataset for training, reflective of the latest environmental interactions.

These tailored strategies highlight a nuanced understanding of how different indicators behave and affect the learning process, ensuring that both dynamic and static aspects of the training data are optimally utilized for updating the diffuser.

\subsection{State-level Trajectory Generation}
In the state-level generation, we directly generate trajectories consisting of states, actions and rewards, $\{s_t, a_t, r_t, s_{t+1}, a_{t+1}, r_{t+1}\dots\}$. For the architecture of the diffusion model used in the state-level generation, we directly refer to the architecture used in \citet{lu2023synthetic}. Meanwhile, we extend the size of the network used from $24$ to $128$ to support the increased length of generated trajectories.

\subsection{Image Encoder}

In the image-level generation, an encoder and a decoder are employed for image rendering and key point detection tasks, respectively. The encoder, based on previous work \citep{vd4rl, modem, DBLP:conf/corl/SeoHLLJLA22}, captures observations in an environment where an agent operates. The rendered images, set at a resolution of $84 \times 84$, serve as input of the generation model.

\section{Details of Experiment Setup}

\subsection{D4RL}

We consider three different environments from D4RL~\citep{fu2020d4rl}. We use the original offline dataset from D4RL~\citep{fu2020d4rl}.

\textbf{AntMaze}. This domain steps up the complexity by replacing the 2D ball in Maze2D with an 8-DoF ``Ant'' quadruped robot, adding a layer of morphological complexity. It is a navigation domain that closely resembles real-world robotic navigation tasks. We follow the design of a sparse 0-1 reward strategy in this environment, activated only upon reaching the goal, to test the stitching challenge under more complex conditions.

\textbf{Locomotion}. Comprising tasks like Hopper, HalfCheetah, and Walker2d, the Locomotion domain is a staple in offline deep RL benchmarks. We use the same datasets in D4RL~\citep{fu2020d4rl} for consistency with previous studies, and also experiment with a variety of datasets to observe the effects of different data qualities.

\textbf{Kitchen}. This environment involves controlling a 9-DoF Franka robot in a kitchen setting, interacting with everyday household items like a microwave, kettle, cabinets, an overhead light, and an oven. Each task aims to achieve a specific goal configuration, like opening the microwave and sliding cabinet door, placing the kettle on the burner, and turning on the overhead light. The Kitchen domain serves as a benchmark for multi-task behavior in a realistic, non-navigation setting. Here, the stitching challenge is amplified due to the complexity of the trajectories through the state space, compelling algorithms to generalize to unseen states rather than rely solely on training trajectories.

\subsection{Meta-World}

The Meta-World benchmark~\citep{DBLP:conf/corl/YuQHJHFL19} is a comprehensive suite designed for evaluating and advancing reinforcement learning and multi-task learning algorithms. It features 50 distinct robotic manipulation tasks, offering a diverse and challenging environment for testing the ability of algorithms to generalize and quickly acquire new skills. Following previous work~\citep{modem}, we select a total of 15 tasks from Meta-World based on their difficulty according to~\citet{DBLP:conf/corl/SeoHLLJLA22}, which categorizes tasks into \emph{easy}, \emph{medium}, \emph{hard}, and \emph{very hard} categories. Same as~\citet{modem}, we discard \emph{easy} tasks and select all tasks from the remaining 3 categories. Our approach differs from previous studies~\citep{modem, DBLP:conf/corl/SeoHLLJLA22} in that we solely utilize proprioceptive state information as input of RL agents, rather than RGB frames or combined state information. This choice allows for easier application to general reinforcement learning methods.  Following the previous settings, we use a sparse reward signal that only provides a reward of $1$ when the current task is solved and $0$ otherwise. For the success criteria, we follow the original setting in Meta-World. Table~\ref{tab:metaworld-tasks} details our multi-task and sequence-task settings.

\textbf{Training Dataset}. The dataset is acquired by training an RL agent with SAC~\citep{sac} on each single task. For the multi-task setting, the entire dataset is combined by all 14 single tasks. For the sequence task, the entire dataset is combined by all 8 single tasks, without any trajectory change.

\begin{table}
\centering
\parbox{\textwidth}{
\centering
\resizebox{\textwidth}{!}{%
\begin{tabular}{@{}ll@{}}
\toprule
Task Name                                                                  & Subtasks                                                                             \\ \midrule
\texttt{Multitask}                                                        & \texttt{evaluation task}: Push                                                                                      \\
\texttt{Sequencetask}                                                  &   Sweep, Sweep Into, Coffee Push, Box Close, Push Wall, Peg Insert Side,  Basketball, Soccer          
\\ \bottomrule                                                                            \\
\end{tabular}%
}
}

\caption{Meta-World task settings. We select the push task as the evaluation task for the multi-task environment. We arrange all the $8$ \emph{medium} tasks to form the sequence-task environment.}
\label{tab:metaworld-tasks}
\end{table}

\section{Additional Experimental Results}

This section provides supplementary experimental findings, offering detailed insights and further analysis to deepen understanding of the properties of ATraDiff and its components.

\subsection{Detailed Online RL Results}
\label{appendix:full_result}

\begin{figure}[htb]
    \centering
    \includegraphics[width=\linewidth]{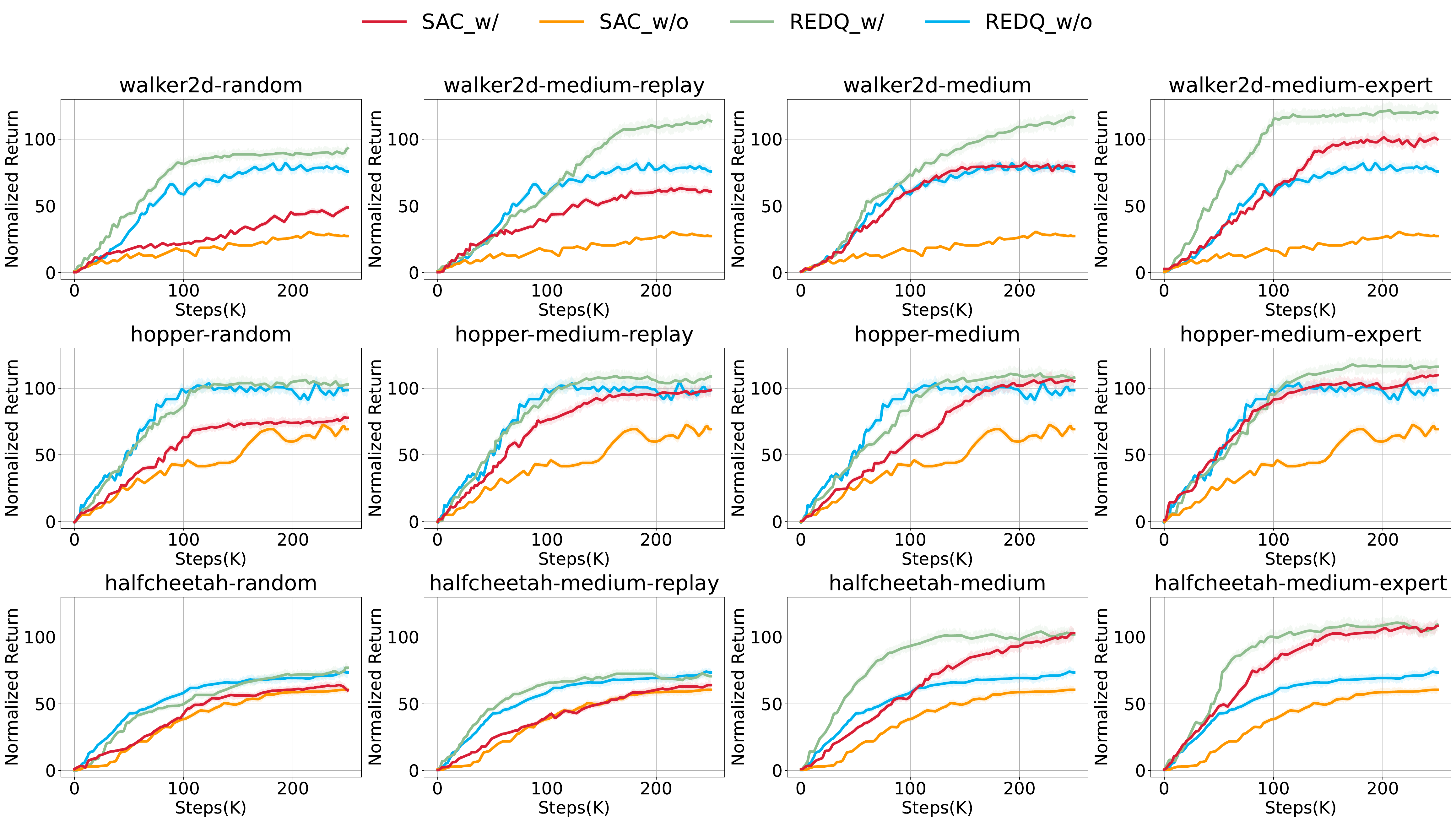}
    \caption{Learning curves of online RL on the D4RL Locomotion benchmark. ATraDiff (denoted as `w/') is comparable to or better than the original RL baseline methods with low-quality data, and the performance gap between ATraDiff and the baselines becomes much more pronounced with medium/high-quality data.}
    \label{figure:full}    
\end{figure}

Here we present the detailed online RL results of each environment on D4RL Locomotion with all 4 different offline datasets. Figure~\ref{figure:full} shows that our ATraDiff is comparable to or better than the original RL baseline methods with low-quality data (such as random data), and the performance gap between our ATraDiff and the baselines becomes much more pronounced with medium/high-quality data (such as medium-expert data).

\subsection{Training Time}
\label{sec:time}

The experiments are conducted on a single NVIDIA RTX 4090TI GPU. The training time of ATraDiff is listed in Table~\ref{table:training_time}.

\begin{table}[h!]
    \centering
    \resizebox{.55\textwidth}{!}{
    \begin{tabular}{c|c|c|c}
      \hline
       &  Walker2D & Hopper & Halfcheetah \\ \hline
      SAC~\citep{sac} & 6.5h & 7h  & 7h \\ \hline
      SAC with ATraDiff & 15h & 17h & 16.5h \\ \hline
      RDEQ~\citep{redq} & 6h & 6.5h & 6.5h \\ \hline
      RDEQ with ATraDiff & 14.5h & 16.5h & 16h \\ \hline
    \end{tabular}
    }
    \caption{Time cost (hours) of training SAC and RDEQ with and without ATraDiff.}
    \label{table:training_time}
\end{table}

\subsection{Comprehensive Analysis and Further Comparison of ATraDiff and SynthER}
\label{appendix:atra_syn}

ATraDiff differs significantly from SynthER~\citep{lu2023synthetic} in various crucial aspects, particularly in the context of the online learning environment. These aspects collectively contribute to our substantial improvements over SynthER. Below, we summarize several key differences.

\begin{itemize}
    \item ATraDiff can perform both image and state generation, with our findings indicating a superior performance of image generation within our framework. This enhancement is likely attributable to the more effective utilization of the pretrained generative model in image generation, as opposed to SynthER's state-only generation approach.
    \item Moreover, ATraDiff is designed to synthesize entire trajectories rather than generating individual transitions as in SynthER. We posit that this capability to produce full trajectories yields a richer, more informative dataset for the learning algorithm than isolated transitions.
    \item In the realm of online training, ATraDiff uniquely generates these trajectories based on the specific current state and task, thereby optimizing the relevance and quality of the generated data for immediate online learning applications.
    \item In addition to these features, during the online adaptation phase, ATraDiff employs importance sampling strategies to selectively leverage data collected online, prioritizing information that is most beneficial for the online adaptation.
\end{itemize}

To substantiate the advantages of ATraDiff, we conduct an addtional ablation study that systematically evaluates the incremental contributions of its components mentioned above. The study is structured as follows: initially, compared with SynthER, ATraDiff is applied to generate pixel-level single transitions without auxiliary mechanisms, denoted as ``\_transition.'' Subsequently, we evaluate the generation of pixel-level trajectories, still devoid of supplementary techniques, denoted as ``\_trajectory.'' Next, we assessed ATraDiff's capability to produce pixel-level trajectories informed by the current state, albeit without implementing importance sampling, denoted as ``\_condition.'' Lastly, we evaluate the full ATraDiff, denoted as ``\_ATraDiff.'' The results from this experiment, as presented in Table~\ref{table:analy}, demonstrate ATraDiff's progressive improvement, affirming that each technique's integration significantly boosts the model's effectiveness, particularly under online training conditions. These findings underscore ATraDiff's novel, sophisticated methodology in enhancing learning and adaptation in dynamic environments, and explain why ATraDiff notably outperforms SynthER.

\begin{table}[h!]
    \centering
    \resizebox{.85\textwidth}{!}{
    \begin{tabular}{c|ccccc}
      \hline
      
\textbf{Task Name} &	\textbf{SynthER}	 & \textbf{RLPD\_transition}	& \textbf{RLPD\_trajectory} &	\textbf{RLPD\_condition} &	\textbf{RLPD\_ATraDiff} \\ \hline
Locomotion &	94.6 &	93.5 &	97.1 &	101.3 &	102.3 \\ \hline
AntMaze &	97.3 &	97.2 &	97.6 &	98.2	& 98.7 \\ \hline
Multitask &	0.57 &	0.60 &	0.72 &	0.75 &	0.77 \\ \hline
Sequencetask &	0.62 &	0.64 &	0.71 &	0.79 &	0.81 \\ \hline
    \end{tabular}
    }
    \caption{Full comparison between ATraDiff and SynthER on both D4RL (normalized return) and Meta-World (return) benchmarks. ATraDiff notably outperforms SynthER, with each of our proposed components contributing to the improvement.}
    \label{table:analy}
\end{table}

\section{Additional Related Work: Discussions on World Models in RL}

As an emerging research direction, we recognize the significant potential of world models in decision-making domains. However, we would like to clarify that our work aims to demonstrate a specific application of world models within the RL framework, which is novel, under-explored, and effective, but we do not claim that our method represents the best way of exploiting world models in RL. An evaluation of the optimal strategies is beyond the scope of this paper, and we leave it as interesting future work. We also believe that the optimal approach will likely vary depending on the application and task at hand.

Distinct from studies that create comprehensive decision-making models capable of autonomously completing tasks (e.g., SUSIE~\citep{SUISE} and Dreamerv2~\citep{dreamerv2}), our work focuses on developing a \textit{versatile auxiliary} method designed to enhance existing RL algorithms. Our contribution is not a standalone task-solving mechanism; rather, it is a plug-in approach intended to integrate seamlessly with any RL algorithm that employs a replay buffer. This distinction underscores the innovative aspect of our work: offering a method that broadens the applicability and efficiency of world models as supportive tools in the RL landscape, thereby facilitating their adoption and utility in diverse RL contexts and applications.

\section{Visualizations}
\label{appendix:vis}

\begin{figure}[h]
    \centering
    \begin{tabular}{@{} l @{\hskip 2mm} c @{\hskip 2mm} c @{\hskip 2mm} c @{\hskip 2mm} c @{\hskip 2mm} c @{\hskip 2mm} c}
        & \textbf{initial} & \textbf{t=0} & \textbf{t=5} & \textbf{t=10} & \textbf{t=15} & \textbf{t=20} \\
        \rotatebox{90}{\footnotesize Sweep} &
        \includegraphics[width=0.15\textwidth]{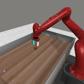} &
        \includegraphics[width=0.15\textwidth]{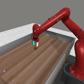} &
        \includegraphics[width=0.15\textwidth]{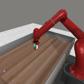} &
        \includegraphics[width=0.15\textwidth]{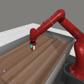} &
        \includegraphics[width=0.15\textwidth]{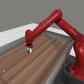} &
        \includegraphics[width=0.15\textwidth]{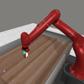} \\
        \rotatebox{90}{\footnotesize Sweep} &
        \includegraphics[width=0.15\textwidth]{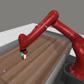} &
        \includegraphics[width=0.15\textwidth]{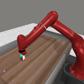} &
        \includegraphics[width=0.15\textwidth]{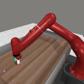} &
        \includegraphics[width=0.15\textwidth]{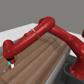} &
        \includegraphics[width=0.15\textwidth]{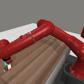} &
        \includegraphics[width=0.15\textwidth]{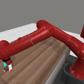} \\
        \rotatebox{90}{\footnotesize Push Wall} &
        \includegraphics[width=0.15\textwidth]{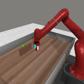} &
        \includegraphics[width=0.15\textwidth]{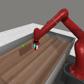} &
        \includegraphics[width=0.15\textwidth]{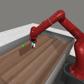} &
        \includegraphics[width=0.15\textwidth]{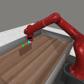} &
        \includegraphics[width=0.15\textwidth]{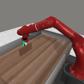} &
        \includegraphics[width=0.15\textwidth]{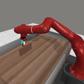} \\
        \rotatebox{90}{\footnotesize Soccer} &
        \includegraphics[width=0.15\textwidth]{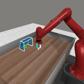} &
        \includegraphics[width=0.15\textwidth]{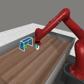} &
        \includegraphics[width=0.15\textwidth]{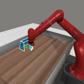} &
        \includegraphics[width=0.15\textwidth]{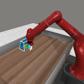} &
        \includegraphics[width=0.15\textwidth]{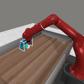} &
        \includegraphics[width=0.15\textwidth]{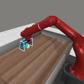} \\
    \end{tabular}
    \caption{Representative visualizations of our generated image trajectories. Each row contains one generation, the initial image represents the state received by ATraDiff, and the following images represent the generated results in different time-steps.}
    \label{fig:appendix-demos}
\end{figure}

In Figure~\ref{fig:appendix-demos}, we show some representative visualizations of our generated image trajectories. We present a variety of images generated under different conditions, including varying initial states and a range of tasks. The labels in the left of the figures represent the task. The results demonstrate the robustness and effectiveness of our image generation method, consistently performing well across diverse scenarios and producing high-quality and temporally coherent image trajectories.

\end{document}